\pdfoutput=1

\documentclass[11pt]{article}

\usepackage[preprint]{acl}

\usepackage{times}
\usepackage{latexsym}
\usepackage{amsmath}
\usepackage{amsmath, amssymb}
\usepackage{algorithm}
\usepackage{listings}
\usepackage{algpseudocode}
\usepackage{booktabs}
\usepackage{multirow}
\usepackage{arydshln}
\usepackage[T1]{fontenc}

\usepackage[utf8]{inputenc}

\usepackage{microtype}

\usepackage{inconsolata}

\usepackage{graphicx}
\usepackage{booktabs}
\usepackage{multirow}
\usepackage{adjustbox}

\newcommand{\tocite}[1]{{\bf \color{blue}{[tocite]}}}
\title{All-in-One Tuning and Structural Pruning for Domain-Specific LLMs}

\author{
 \textbf{Lei Lu\textsuperscript{1}},
 \textbf{Zhepeng Wang\textsuperscript{2}},
 \textbf{Runxue Bao\textsuperscript{3}},
 \textbf{Mengbing Wang\textsuperscript{1}},
 \textbf{Fangyi Li\textsuperscript{4}}, \\
 \textbf{Yawen Wu\textsuperscript{5}},
 \textbf{Weiwen Jiang\textsuperscript{2}},
 \textbf{Jie Xu\textsuperscript{6}},
 \textbf{Yanzhi Wang\textsuperscript{1}}, 
 \textbf{Shangqian Gao\textsuperscript{7}}
\\
 \textsuperscript{1}Northeastern University,
 \textsuperscript{2}George Mason University, 
 \textsuperscript{3}GE HealthCare, 
 \textsuperscript{4}University of Pennsylvania, \\
 \textsuperscript{5}University of Pittsburgh,
 \textsuperscript{6}University of Florida,
 \textsuperscript{7}Florida State University\\
\textsuperscript{1}\texttt{\{lu.lei1, wang.mengb, yanz.wang\}@northeastern.edu}, \\
\textsuperscript{2}\texttt{\{zwang48@, wjiang8\}@gmu.edu},
\textsuperscript{3}\texttt{runxue.bao@gehealthcare.com},\\ 
\textsuperscript{4}\texttt{fangyili@seas.upenn.edu}, 
\textsuperscript{5}\texttt{yawen.wu@pitt.edu}, 
\textsuperscript{6}\texttt{xujie@ufl.edu},
\textsuperscript{7}\texttt{sgao@cs.fsu.edu}
\\
}

\begin{document}
\maketitle
\begin{abstract}
Existing pruning techniques for large language models (LLMs) targeting domain-specific applications typically follow a two-stage process: pruning the pretrained general-purpose LLMs and then fine-tuning the pruned LLMs on specific domains. However, the pruning decisions, derived from the pretrained weights, remain unchanged during fine-tuning, even if the weights have been updated. Therefore, such a combination of the pruning decisions and the finetuned weights may be suboptimal, leading to non-negligible performance degradation.
To address these limitations, we propose \textbf{ATP}: \textbf{A}ll-in-One \textbf{T}uning and Structural \textbf{P}runing, a unified one-stage structural pruning and fine-tuning approach that dynamically identifies the current optimal substructure throughout the fine-tuning phase via a trainable pruning decision generator. Moreover, given the limited available data for domain-specific applications, Low-Rank Adaptation (LoRA) becomes a common technique to fine-tune the LLMs. In ATP, we introduce LoRA-aware forward and sparsity regularization to ensure that the substructures corresponding to the learned pruning decisions can be directly removed after the ATP process. ATP outperforms the state-of-the-art two-stage pruning methods on tasks in the legal and healthcare domains. More specifically, ATP recovers up to 88\% and 91\% performance of the dense model when pruning 40\% parameters of LLaMA2-7B and LLaMA3-8B models, respectively.


\end{abstract}

\section{Introduction}

Domain-specific LLMs have become indispensable for handling professional tasks such as legal, healthcare, and finance applications~\cite{ling2024domainspecializationkeymake,jeong2024fine,zheng2023trafficsafetygpt}. By fine-tuning the general-purpose pretrained LLMs~\cite{wang2024infuserki,wang2024self,wang2024unlocking} on domain-specific datasets~\cite{zheng2024fine,susnjak2024automating,xie2024me}, the domain-specific LLMs can adapt to the unique terminologies and nuanced contextual requirements of the given domain, producing high-quality outputs of the relevant domain.

Due to the limited size of domain-specific datasets, full-parameter fine-tuning of LLMs is usually prone to significant knowledge forgetting and performance degradation~\cite{christophe2024med42,wang2024infuserki,lin2024data}. To mitigate this issue, Parameter-Efficient Fine-Tuning (PEFT) methods, such as Low-Rank Adaptation (LoRA)~\cite{hu2021lora}, have been widely adopted. These techniques enable effective domain alignment while retaining the knowledge and capabilities of the general-purpose LLMs to the greatest extent. However, domain-specific LLMs inherit the substantial computational and memory costs of their general-purpose counterparts, posing significant challenges for deployment \cite{wan2023efficient, stojkovic2024towards}. Therefore, effective compression techniques are important to deploy the LLMs onto domain applications. 

\begin{figure}[t]
    \centering
    \includegraphics[width=\linewidth]{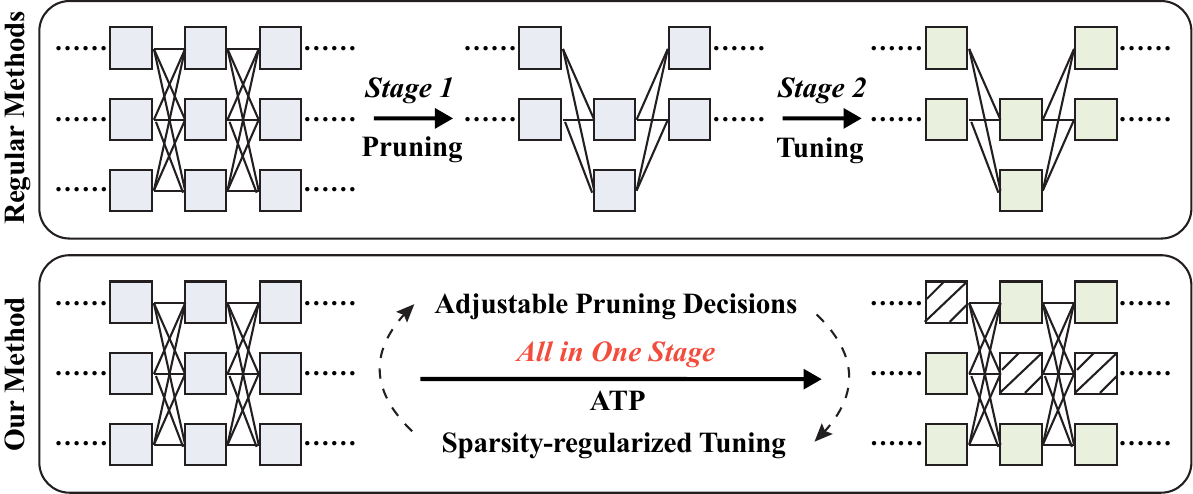}
    \caption{Comparison between regular pruning methods and our method (ATP). Our one-stage design unifies regular two-stage methods while outperforming them.}
    \label{fig:prunable_weights}
    \vspace{-10pt}
\end{figure}

Pruning is one of the most promising solutions to compress the LLMs by systematically removing less significant parameters~\cite{xia2023sheared, guo2023compresso, gao2024disp, an2024fluctuation}. To construct compact models for cost-efficient deployment, structural pruning is usually preferable to weight-level pruning. Although structural pruning can effectively reduce computational and memory overhead, it often incurs non-negligible performance degradation, especially when the sparsity level is high (e.g., $\geq 0.4$). To address it, current pruning techniques typically follow a two-stage pipeline: (1) pruning the general-purpose pretrained model to produce a compact model, and (2) fine-tuning the pruned model, aiming to recover the performance loss incurred by pruning while aligning the model with domain knowledge. The major issue of this pipeline is that the pruning decisions obtained from the pretrained weights during the pruning stage remain unchanged during the fine-tuning stage. However, the optimal substructure of the pretrained model may evolve during fine-tuning since the importance of weights can change when the weights are kept updated. Ignoring such structural evolution is likely to result in sub-optimal pruning decisions for domain-specific applications.

In this paper, we propose \textbf{ATP}, \textbf{A}ll-in-One \textbf{T}uning and \textbf{P}runing, a unified one-stage framework for domain-specific structural pruning of LLMs that integrates pruning-decision search and LoRA-based fine-tuning, targeting for the domain application where limited fine-tuning data is available. ATP establishes a dynamic interplay between structural pruning and parameter fine-tuning by continuously updating the pruning decision everytime the weight is updated while constraining the fine-tuning of weights with the current pruning decision simultaneously. More specifically, a pruning-decision generator is introduced to continuously generate pruning decisions based on the updated weights, enabling the exploration of optimal substructures throughout the tuning process. The effect of the current pruning decision on the model output is simulated through a LoRA-aware forward pass without actual model compression. Concurrently, the LoRA-aware structural sparsity constraints are proposed to penalize the LoRA weights associated with the pruned structures indicated by the pruning decisions. The penalization gradually diminishes the contribution of the relevant weights to the final output, such that both the LoRA weights and their corresponding pretrained weights can be effectively removed at the end. The output of ATP is a compact model tailored to the given domain, derived from the final pruning decisions.

We evaluated ATP in the \textit{HealthCare} and \textit{Legal} domains. Experimental results demonstrate that ATP outperforms conventional two-stage structural pruning in both language modeling and problem-solving capabilities across different domains. Notably, even at high sparsity levels ($\geq 0.4$), the performance of models pruned with ATP remains comparable to that of the domain-specific counterpart without pruning.

Our contributions are summarized as follows:
\begin{itemize} 
\item We propose ATP, a one-stage approach that integrates structural pruning with LoRA-based fine-tuning, optimized for domain-specific applications with limited fine-tuning data.
    

\item We design a novel sparsity-constrained tuning method tailored for LoRA by introducing a LoRA-aware forward pass and structural sparsity regularization to LoRA weights.
    
\item We conduct extensive experiments in the \textit{HealthCare} and \textit{Legal} domains. The results show that ATP outperforms the conventional two-stage structural pruning in domain-specific applications in most of the evaluated settings. More specifically, it recovers up to 88\% and 91\% performance of the dense model when pruning 40\% parameters of LLaMA2-7B and LLaMA3-8B, respectively.
\end{itemize}

\section{Related Work}
\noindent\textbf{LLM Pruning.} To reduce the computational cost of machine learning algorithms, model pruning was proposed and has achieved great success in the training and inference stage of conventional machine learning models \cite{bao2020fast, bao2022accelerated, bao2022doubly} and convolutional neural networks (CNNs)~\cite{wang2021lightweight, wu2020enabling, wu2020intermittent, wang2023edge}. Therefore, applying the pruning techniques to LLMs seems to be an intuitive and promising method. However, the new architecture from transformer layers, the huge amount of parameters and the higher expectation of the capability of the pruned model pose new challenges to the pruning of LLMs and the effective method to prune LLMs is still under exploration. Existing pruning methods for LLMs can be classified into unstructured and structured pruning. Unstructured pruning removes individual weights, resulting in a sparse model that maintains the original structure. In contrast, structured pruning eliminates entire channels or layers, producing a smaller model with reduced dimensionality.

Unstructured pruning methods \cite{frantar2023sparsegpt,sun2024simpleeffectivepruningapproach, zhang2023loraprune, zhang2024pruning} have shown promising results for general and domain-specific applications, retaining comparable performance with the dense model. Structured pruning further enhances deployment compatibility by removing groups of weights, but identifying optimal pruning patterns while preserving performance is more challenging.

Recent structural pruning techniques focus on finding optimal pruning patterns. For example, \citet{ma2023llm} calculates grouped importance scores to eliminate less significant components, \citet{lin2024modegpt} utilizes combined matrix decomposition to determine pruning pattern without propagation, \citet{ashkboos2023slicegpt} uses orthogonal transformations for matrix slicing, and \citet{van2023llm} employs multi-step search strategies to optimize pruning patterns. Most of them typically incorporate an additional post-pruning fine-tuning stage for performance recovery at higher sparsity levels. Such separation prevents mutual interaction between pruning and fine-tuning, failing to consider the change in the optimal pruning pattern due to weight updates \cite{wu2024auto}. Hence, bridging such a connection between pruning decisions and tuning is the main focus of ATP.

\noindent\textbf{LLM Fine-Tuning.} Pretraining a representation model via self-supervised learning~\cite{wu2021enabling,wu2021federated} and then fine-tuning the pretrained model to downstream tasks follows the principles of transfer learning, which have been extensively used in computer vision. Transfer learning enables leveraging knowledge gained from a source domain (e.g., a large-scale, general-purpose corpus) to enhance performance in a target domain (e.g., domain-specific datasets)~\cite{zhang2024neurodegenerative,bao2024recentsurveyheterogeneoustransfer}. This approach reduces the training cost for domain-specific applications compared with training the model from scratch~\cite{wu2021decentralized,wu2022distributed,wu2021federated2,zhang2022toward,wu2023synthetic}. In the development of large language models (LLMs), a similar transfer learning paradigm is adopted. The LLM is first pretrained on a massive general-purpose corpus and then fine-tuned on small domain-specific datasets to align the pretrained general models toward specialized domains. Full-parameter fine-tuning of LLMs~\cite{lv2023full} updates all model parameters, but it presents challenges like high computational demands and the risk of over-fitting, especially with limited data \cite{zhang2024scaling}.

Transfer learning-based PEFT methods~\cite{houlsby2019parameter,he2021towards,lester2021power,liu2021p,hu2021lora,liu2024dora,wang2024infuserki} have been thus developed to address these challenges, which generally adapt LLMs without model weight update via prompt tuning and pre-fix tuning \cite{li2021prefixtuningoptimizingcontinuousprompts}, or train only a small subset of parameters while keeping the rest of the model frozen with extra modules. Among these, LoRA~\cite{hu2021lora} stands out as a representative method for PEFT due to its efficiency and compatibility. LoRA inserts trainable low-rank matrices into each Transformer layer, allowing adaptation with minimal additional parameters. This reduces computational and memory overhead and helps preserve the generalization of the LLM after parameter tuning. In this work, we build the foundation of ATP based on LoRA-tuning to satisfy the usually limited fine-tuning data within a specific domain.

\section{Methodology}
ATP dynamically searches for the optimal pruning decision via a trainable pruning-decision generator alongside the LoRA-tuning process. Upon convergence, the LoRA-weights corresponding to the pruned structures approach zero, allowing for the direct extraction of a compressed and fine-tuned domain-specific LLM guided by the finalized pruning decision.

\begin{figure*}[ht]
    \centering
    \includegraphics[width=0.9\linewidth]{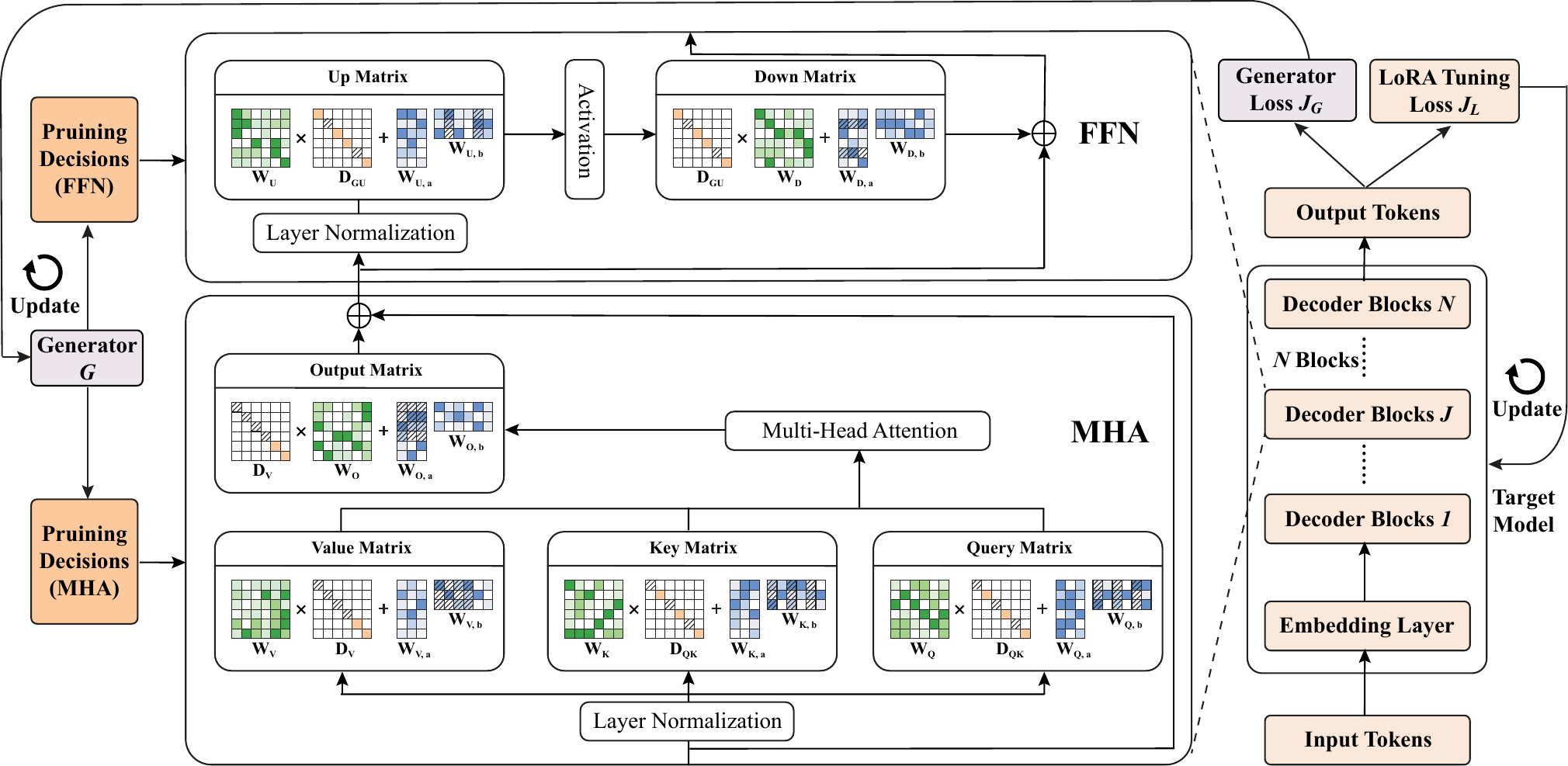}
    \caption{Overview of one training step of ATP.}
    \label{fig:pruning_pipeline}
    \vspace{-10pt}
\end{figure*}

\subsection{Notations}
To clarify our methodology, we define the following notations. For a linear projection in LLMs, let $\mathbf{W} \in \mathbb{R}^{m \times n}$ be the pretrained weight matrix, and $\mathbf{X} \in \mathbb{R}^{l \times m}$ be the input feature. The corresponding LoRA module's weights are denoted as $\mathbf{W_a} \in \mathbb{R}^{m \times r}$ and $\mathbf{W_b} \in \mathbb{R}^{r \times n}$, where $m$ and $n$ are the input and output dimensions, respectively, $l$ is the number of tokens, and $r$ is the LoRA rank. Let $\mathbf{D} \in \mathbb{R}^{n \times n}$ be a diagonal pruning-selection matrix with entries in $\{0, 1\}$, satisfying $\mathbf{D}^2 = \mathbf{D}$.  We further denote the diagonal vector of $\mathbf{D}$ as the pruning decision vector $\mathbf{d} \in \{0, 1\}^{n}$, where $d_i = 1$ signifies that the $i$-th output dimension is retained, and $d_i = 0$ indicates pruning. If the output dimension remains unpruned, then $\mathbf{D} = \mathbf{I}$, where $\mathbf{I} \in \mathbb{R}^{n \times n}$ is the identity matrix.

\vspace{-5pt}
\subsection{Prunable Groups in LLMs}
We perform structural pruning in an LLM by group-wise removal on linear projections within decoder layers. Specifically, each row and column of \(\mathbf{W}\) is treated as an individual prunable group.  

A decoder layer typically consists of two sequential blocks: Attention and Multi-Layer Perceptron (MLP). Let $d_h$, $d_{int}$ and $H$ denote the hidden dimension, intermediate dimension, and the number of attention heads, respectively, pruning these blocks can be mathematically expressed as follows:
\begin{equation}
\begin{aligned}
    f_{\text{MLP}}(\mathbf{X}) &= \left( \mathcal{A}(\mathbf{X} \mathbf{W}_{\text{\tiny G}} \mathbf{D}_{\text{\tiny GU}}) 
    \odot (\mathbf{X} \mathbf{W}_{\text{\tiny U}} \mathbf{D}_{\text{\tiny GU}}) \right) \mathbf{W}_{\text{\tiny D}}, \\
    f_{\text{Attn}}(\mathbf{X}) &= \text{Concat}(\text{head}_1, \dots, \text{head}_h) \mathbf{W}_{\text{\tiny O}}, \\
    \text{head}_i &= \text{softmax}\left( \mathbf{Q}_i \mathbf{K}_i^\top \right) \mathbf{V}_i.
\end{aligned}
\label{eq:mlp_attn_prune}
\end{equation}
where \(\mathcal{A}\) denotes the activation function, \(\odot\) indicates element-wise multiplication, and \(\mathbf{W}_{\text{\tiny G}}, \mathbf{W}_{\text{\tiny U}} \in \mathbb{R}^{d_h \times d_{int}}, \mathbf{W}_{\text{\tiny D}} \in \mathbb{R}^{d_{int} \times d_h}\) represent the gate, up, and down weight matrices of the MLP block. In the attention block, the query, key, and value \(\mathbf{Q}_i\), \(\mathbf{K}_i\), and \(\mathbf{V}_i\) for $\text{head}_{i}$ are defined respectively as:
$\mathbf{Q}_i = \mathbf{X} \mathbf{W}_{\text{\tiny Q}_i}  \mathbf{D}_{\text{\tiny QK}}$, 
$\mathbf{K}_i = \mathbf{X} \mathbf{W}_{\text{\tiny K}_i}  \mathbf{D}_{\text{\tiny QK}}$, 
$\mathbf{V}_i = \mathbf{X} \mathbf{W}_{\text{\tiny V}_i}  \mathbf{D}_{\text{\tiny V}}$, 
where $\mathbf{W}_{\text{\tiny Q}_i}, \mathbf{W}_{\text{\tiny K}_i}, \mathbf{W}_{\text{\tiny V}_i}  \in \mathbb{R}^{d_h \times \frac{d_h}{H}}$ are the head-specific weight matrices for the query, key, and value, $\mathbf{W}_{\text{\tiny O}}$ is the weight matrix for out, $\mathbf{D}_{\text{\tiny QK}}, \mathbf{D}_{\text{\tiny V}} \in \mathbf{R}^{ \frac{d_h}{H} \times \frac{d_h}{H}}$. 

As shown in Eq.\ref{eq:mlp_attn_prune}, we keep the output dimensions of $\mathbf{W}_{\text{\tiny D}}$ and $\mathbf{W}_{\text{\tiny O}}$ unpruned to ensure dimensional consistency of residual connections across layers.
To ensure the same head dimension across all heads, the same pruning-selection matrix $\mathbf{D}_{\text{\tiny QK}}$ and $\mathbf{D}_{\text{\tiny V}}$ is applied to every head. Furthermore, \(\mathbf{D}_{\text{\tiny V}}\) indirectly prunes the input rows of \(\mathbf{W}_{\text{\tiny O}}\), while \(\mathbf{D}_{\text{\tiny GU}}\) similarly prunes the input rows of \(\mathbf{W}_{\text{\tiny D}}\), due to the adjacency of their respective linear projections.

Thus, our pruning design simplifies to searching for the optimal \(\mathbf{D}_{\text{\tiny QK}}\), \(\mathbf{D}_{\text{\tiny V}}\), and \(\mathbf{D}_{\text{\tiny GU}}\) for each decoder layer, where these pruning decisions select a subset from the prunable groups within an LLM to form the final removal set \(\mathcal{G}\).

\vspace{-5pt}

\subsection{Pruning-Decision Generator}
To dynamically generate optimal pruning decisions, we introduce a trainable pruning-decision generator \(\mathbf{G}\), which outputs \(\mathbf{d}_{all} = \{\mathbf{d}_{1}, \cdots, \mathbf{d}_{n}, \cdots, \mathbf{d}_{N}\}\), a set of \(N\) pruning-decision vectors corresponding to each of the \(N\) decoder layers. $\mathbf{d}_{n}$ for $n$-th layer is the  concatenation of \(\mathbf{d}_{\text{\tiny QK}}\), \(\mathbf{d}_{\text{\tiny V}}\), and \(\mathbf{d}_{\text{\tiny GU}}\).

The generator \(\mathbf{G}\) is constructed sequentially with Transformer encoder blocks, followed by fully connected layers that project the output to the dimension of \(len(\mathbf{d}_n)\). The Gumbel-Sigmoid~\cite{jang2016categorical} function, combined with the straight-through estimator (STE), serves as the final output layer to produce decision vectors that closely approximate a binomial distribution. Given trainable weights \(\mathbf{M}\) of $\mathbf{G}$, the generator produces the set of pruning-decision vectors as:
\begin{equation}
    \mathbf{d}_{\text{all}} = \mathbf{G}(\mathbf{M}).
\vspace{-2pt}
\end{equation}
Such design ensures discrete decision generation while maintaining differentiability. The detailed structure of $\mathbf{G}$ is shown in the Appendix~\ref{sec:a-1}.

\subsection{LoRA-Aware Designs}
We integrate the generated pruning decisions with LoRA-based tuning through two key designs: (1) LoRA-aware forward pass and (2) LoRA-aware sparsity regularization.

\noindent\textbf{Forward pass for training $\mathbf{G}$.}
Training \(\mathbf{G}\) aims to search for optimal pruning decisions, making it crucial to simulate the actual pruning effect of the decisions on the output behavior of the LoRA-integrated LLM. To achieve this, we formulate the forward pass of a LoRA-linear projection as:
\begin{equation}
    f_{G}(\mathbf{X}) = \mathbf{X} \left( \mathbf{W} + \mathbf{W}_a \mathbf{W}_b \right) \mathbf{D},
\label{eq:eval}
\end{equation}
where the pruned portions of both the pretrained weights \(\mathbf{W}\) and the LoRA module \(\mathbf{W}_a \mathbf{W}_b\) are ignored based on the current pruning decisions.

\noindent\textbf{Forward pass for LoRA weights tuning.} However, during LoRA-based tuning, directly applying \(\mathbf{D}\) disrupts gradient flow to the currently pruned dimensions, preventing updates to the corresponding LoRA parameters. The update stagnation of LoRA weights narrows the search space of \(\mathbf{D}\), ignoring potentially better pruning decisions.
To address this, we formulate the forward pass for LoRA-Linear training as:
\begin{equation}
    f_{\text{L}}(\mathbf{X}) = \mathbf{X} \left( \mathbf{W} \mathbf{D} + \mathbf{W}_a \mathbf{W}_b \right),
\label{eq:train}
\end{equation}
where the pretrained weights are temporarily masked by \(\mathbf{D}\), while the LoRA parameters remain fully trainable.

\noindent\textbf{LoRA-aware sparsity regularization.}
As we aim to directly remove \(\mathcal{G}\) after the ATP process with minimal negative effects, we incorporate sparsity regularization on the LoRA weights to ensure that weights groups pruned by $\mathbf{D}$ approaches $0$. Specifically, we apply LoRA-aware group lasso regularization to drive the pruned portions of the LoRA weights toward zero during training.

Specifically, we separately constrain the rows in \((\mathbf{I} - \mathbf{D}_{prev})\mathbf{W}_a\) and the columns in \(\mathbf{W}_b(\mathbf{I} - \mathbf{D})\) at the decided pruned positions to achieve alignment of structural sparsity between LoRA weights and \(\mathbf{W}\), where \(\mathbf{D}_{prev}\) denotes the pruning matrix of the previous layer, consequentially pruning the input dimensions of the current layer. Thus, the LoRA-aware regularization term is defined as:
\begin{equation}
    \mathcal{L}_{gl} = \left( \sum_{g_a \in \mathcal{G}_{a}} \left\| g_a \right\|_2 + \sum_{g_b \in \mathcal{G}_{b}} \left\| g_b \right\|_2 \right),
\end{equation}
where \(\mathcal{G}_{a}\) represents the set of row groups in \((\mathbf{I} - \mathbf{D}_{prev})\mathbf{W}_a\), \(\mathcal{G}_{b}\) represents the set of column groups in \(\mathbf{W}_b(\mathbf{I} - \mathbf{D})\), respectively and $\|\cdot\|_2$ denotes the $L_2$-norm. The unpruned groups are not affected by this sparsity regularization.

By employing LoRA-aware sparsity regularization, we ensure that the model gradually approaches the structural sparsity dictated by the pruning decisions during the LoRA tuning process while preserving the flexibility for decision updates within the ATP process.

\begin{algorithm}[!t]
\caption{ATP Algorithm}\label{alg:atp}
\begin{algorithmic}[1]
\State \textbf{Input:} Target LLM with trainable LoRA weights \(\mathbf{W}_{L}\); 
Domain-specific training dataset \(\mathcal{D}_{t}\) and calibration dataset \(\mathcal{D}_{c}\); 
Total training steps \(T\); End of pruning-decision update training step \(T_{\text{end}}\).

\State \textbf{Initialization:} Build the pruning-decision generator $\mathbf{G}$ with initialized weights $\mathbf{M}$ that each initial $\mathbf{d}_n \in \mathbf{d}_{all}$ is a vector of ones;

\For{$t = 1$ to $T$}
    \State Sample a mini-batch $s_{c}$ from $\mathcal{D}_{c}$;
    \If{$t \leq T_{\text{end}}$}
        \State $\mathbf{M}, \mathbf{d}_{all} \leftarrow \text{Update\_G}(s_{c}, \mathbf{M}, \mathbf{W}_{L})$;
    \EndIf
    \State Sample a mini-batch $s_{t}$ from $\mathcal{D}_{t}$;
    
    \State Compute the LLM Loss with $\mathbf{d}_{all}$: 
    
    $\mathcal{L}_{LM}(\mathbf{W}_{L}) \leftarrow \text{LLM}.f_{L}(s_{t}; \mathbf{d}_{all})$; 

    \State Compute group lasso loss:
    
    $\mathcal{L}_{gl}(\mathbf{W}_L) \leftarrow \text{group\_lasso} (\mathbf{W}_L, \mathbf{d}_{all});$
    
    \State Update $\mathbf{W}_{L}$ with $\nabla_{\mathbf{W}_{L}} (\mathcal{L}_{LM} + \beta\mathcal{L}_{gl});$
    
\EndFor
\State \textbf{Compression:} Directly remove the pruned groups in $\mathcal{G}$ according to $\mathbf{d}_{all}$;

\State \textbf{Output:} A structural-pruned and fine-tuned LLM $\mathbf{L}_{p}$ and the pruning-decision set $\mathbf{d}_{all}$;
\end{algorithmic}
\end{algorithm}
\vspace{-5pt}

\begin{algorithm}[!t]
\caption{Update\_\ensuremath{\mathbf{G}}}\label{alg:update_g}
\begin{algorithmic}[1]
\State \textbf{Inputs:} Target LLM with LoRA weights $\mathbf{W}_{L}$; Pruning-decision Generator $\mathbf{G}$ with weights $\mathbf{M}$; Calibration sample $s_{c}$;
\State $\mathbf{d}_{all} \leftarrow \mathbf{G}(\mathbf{M})$;
\State Compute the LLM loss with $\mathbf{d}_{all}$:  \\ $\mathcal{L}_\text{LM}(\mathbf{M}) \leftarrow \text{LLM}.f_{G}(s_{c}; \mathbf{d}_{all})$;
\State Compute the decision constraint loss: \\ 
$\mathcal{L}_{s}(\mathbf{M}) \leftarrow \text{sparsity\_constraint}(\mathbf{d}_{all})$;
\State Update $\mathbf{M}$ with gradient $\nabla_{\mathbf{M}} (\mathcal{L}_\text{LM} + \alpha \mathcal{L}_{s})$; 
\State Update pruning decision: \\ $\mathbf{d}_{all} \leftarrow \mathbf{G}(\mathbf{M})$;
\State \textbf{Output:} Updated pruning-decision set $\mathbf{d}_{all}$ and weights $\mathbf{M}$ of $\mathbf{G}$;
\end{algorithmic}
\end{algorithm}

\begin{table*}[!t]
\centering
\resizebox{0.9\textwidth}{!}{%
\begin{tabular}{cccccccccccccc}
\hline
\multicolumn{1}{c|}{{ }} &
  \multicolumn{1}{c|}{{ }} &
  \multicolumn{7}{c|}{{HealthCare}} &
  \multicolumn{5}{c}{{Legal}} \\ \cline{3-14} 
\multicolumn{1}{c|}{{ }} &
  \multicolumn{1}{c|}{{ }} &
  {Harrison} &
  {MedNLI} &
  {PubMedQA} &
  \multicolumn{3}{c}{{HQS}} &
  \multicolumn{1}{c|}{{Relative}} &
  {LegalPile} &
  \multicolumn{3}{c}{{BillSum}} &
  {Relative} \\
\multicolumn{1}{c|}{\multirow{-3}{*}{{ Methods}}} &
  \multicolumn{1}{c|}{\multirow{-3}{*}{{Size$\downarrow$}}} &
  {Perplexity$\downarrow$} &
  {Acc.$\uparrow$} &
  {F1$\uparrow$} &
  {R1$\uparrow$} &
  {R2$\uparrow$} &
  {RL$\uparrow$} &
  \multicolumn{1}{c|}{{\%$\uparrow$}} &
  {Perplexity$\downarrow$} &
  {R1$\uparrow$} &
  {R2$\uparrow$} &
  {RL$\uparrow$} &
  {\%$\uparrow$} \\ \hline
\multicolumn{14}{c}{{ \textbf{LLaMA2-7B (p=0.5)}}} \\ \hline
{ Dense} &
  { 6.74B} &
  { 7.33} &
  { 84.87} &
  { 56.38} &
  { 34.24} &
  { 12.79} &
  { 29.83} &
  { -} &
  { 2.47} &
  { 50.8} &
  { 30.07} &
  { 36.28} &
  { -} \\ \hdashline
{ LLM-Pruner} &
  { 4.11B} &
  { 13.67} &
  { 57.31} &
  { 36.55} &
  { 21.72} &
  { 5.47} &
  { 20.5} &
  { 63.55} &
  { 7.97} &
  { 37.43} &
  { 11.96} &
  { 21.75} &
  { 57.8} \\
{ SliceGPT} &
  { 3.73B} &
  { 23.31} &
  { 67.02} &
  { 40.93} &
  { 22.85} &
  { 5.7} &
  { 20.32} &
  { 70.46} &
  { 12.8} &
  { 23.1} &
  { 10.03} &
  { 16.3} &
  { 41.25} \\
{ \textbf{ATP (ours)}} &
  { \textbf{3.52B}} &
  { \textbf{9.53}} &
  { \textbf{70.51}} &
  { \textbf{42.06}} &
  { \textbf{29.66}} &
  { \textbf{10.36}} &
  { \textbf{27.38}} &
  { \textbf{81.38}} &
  { \textbf{3.67}} &
  { \textbf{43.8}} &
  { \textbf{23.12}} &
  { \textbf{30.06}} &
  { \textbf{81.99}} \\ \hline
\multicolumn{14}{c}{{ \textbf{LLaMA2-7B (p=0.4)}}} \\ \hline
{ LLM-Pruner} &
  { 4.68B} &
  {11.21} &
  { 71.15} &
  { 36.81} &
  { 26.9} &
  { 8.57} &
  { 24.23} &
  { 74.91} &
  { 6.48} &
  { 39.5} &
  { 14.86} &
  { 23.88} &
  { 64.33} \\
{ SliceGPT} &
  { 4.50B} &
  { 18.97} &
  { \textbf{72.29}} &
  { 44.04} &
  { 23.69} &
  { 6.68} &
  { 21.08} &
  { 75.77} &
  { 11.21} &
  { 24.05} &
  { 10.68} &
  { 16.7} &
  { 42.96} \\
{ \textbf{ATP (ours)}} &
  { \textbf{4.15B}} &
  { \textbf{8.47}} &
  { 71.52} &
  { \textbf{44.6}} &
  { \textbf{31.33}} &
  { \textbf{11.15}} &
  { \textbf{28.21}} &
  { \textbf{84.82}} &
  { \textbf{2.82}} &
  { \textbf{46.51}} &
  { \textbf{26.75}} &
  { \textbf{33.8}} &
  { \textbf{91.23}} \\ \hline
\multicolumn{14}{c}{{ \textbf{LLaMA3-8B (p=0.5)}}} \\ \hline
{ Dense} &
  { 8.03B} &
  { 8.91} &
  { 85.51} &
  { 54.52} &
  { 33.95} &
  { 13} &
  { 30.09} &
  { -} &
  { 3.49} &
  { 48.35} &
  { 29.1} &
  { 35.58} &
  { \textbf{-}} \\ \hdashline
{ LLM-Pruner} &
  { 5.19B} &
  { 19.98} &
  { 64.84} &
  { 35.75} &
  { 26.49} &
  { 8.85} &
  { 22.74} &
  { 71.76} &
  { 10.23} &
  { 22.36} &
  { 13.2} &
  { 16.64} &
  { 46.12} \\
{ SliceGPT} &
  { 4.81B} &
  { 41.02} &
  {\textbf{75.04}} &
  { 34.69} &
  { 21.64} &
  { 5.33} &
  { 19.36} &
  { 69.25} &
  { 25.88} &
  { 21.4} &
  { 12.38} &
  { 15.95} &
  { 43.88} \\
{ \textbf{ATP (ours)}} &
  { \textbf{4.57B}} &
  { \textbf{13.94}} &
  { 68.57} &
  { \textbf{36.72}} &
  { \textbf{27.91}} &
  { \textbf{9.7}} &
  { \textbf{24.71}} &
  { \textbf{75.73}} &
  { \textbf{4.28}} &
  { \textbf{43.1}} &
  { \textbf{22.6}} &
  { \textbf{28.65}} &
  { \textbf{82.44}} \\ \hline
\multicolumn{14}{c}{{ \textbf{LLaMA3-8B (p=0.4)}}} \\ \hline
{ LLM-Pruner} &
  { 5.79B} &
  {\textbf{11.21}} &
  { 72.15} &
  { 36.81} &
  { 26.9} &
  { 8.57} &
  { 24.23} &
  { 75.71} &
  { 8.52} &
  { 27.27} &
  { 16.91} &
  { 20.49} &
  { 57.37} \\
{ SliceGPT} &
  { 5.43B} &
  { 16.78} &
  { 74.33} &
  {\textbf{47.04}} &
  { 23.69} &
  { 6.68} &
  { 21.08} &
  { 78.98} &
  { 18.49} &
  { 27.3} &
  { 16.73} &
  { 20.13} &
  { 56.84} \\
{ \textbf{ATP (ours)}} &
  { \textbf{5.24B}} &
  { 12.48} &
  { \textbf{75.32}} &
  {43.09} &
  { \textbf{29.86}} &
  { \textbf{11.16}} &
  { \textbf{27.01}} &
  { \textbf{84.99}} &
  { \textbf{4.13}} &
  { \textbf{45.28}} &
  { \textbf{25.45}} &
  { \textbf{30.15}} &
  { \textbf{88.62}} \\ \hline
\end{tabular}%
}
\caption{Overall results in \textit{HealthCare} and \textit{Legal} Domains, best structural-pruned scores of are \textbf{bold} numbers and $p$ is the desired sparsity level. $\uparrow$ indicates the higher better, $\downarrow$ indicates the lower the better. The relative performance of each pruned model is calculated as shown in Appendix \ref{sec:a-2}.}
\label{tab:performance_comparison}
\vspace{-10pt}
\end{table*}

\subsection{ATP Algorithm}
The overall pipeline of ATP is shown in Fig.\ref{fig:pruning_pipeline}. We further introduce the ATP algorithm design, as detailed in Alg.\ref{alg:atp} and Alg.\ref{alg:update_g}.

We first construct the pruning-decision generator \(\mathbf{G}\) according to the configuration of the target LLM. The initial pruning-decision vectors are all ones, indicating no pruning at the start, and are gradually refined to identify optimal pruning decisions.

In each ATP step~(Fig.\ref{fig:pruning_pipeline}), \(\mathbf{G}\) is updated using a small calibration dataset \(\mathcal{D}_{s}\). The updated pruning-decision set \(\mathbf{d}_{all}\) must satisfy two principles: (1) the forward pass affected by pruning decisions of the LLM should preserve the performance on the calibration data, and (2) the overall pruning decisions must follow the desired sparsity level of the LLM. Thus, the optimization objective for \(\mathbf{G}\) can be defined as:
\begin{equation}
\begin{aligned}
\min_{\mathbf{M}} J_{G}(\mathbf{M}) &:= \mathcal{L}_{\text{LM}}(f_G(d_c; \mathbf{d}_{all})) + \alpha \mathcal{L}_{\text{s}}(\mathbf{d}_{all}), \\
\mathcal{L}_{\text{s}}(\mathbf{d}_{all}) &= \text{log}\left(\frac{\max(R(\mathbf{d}_{all}),\ r P_{\text{total}})}{\min(R(\mathbf{d}_{all}),\ r P_{\text{total}})}\right),
\end{aligned}
\label{eq:loss_generator}
\end{equation}
where \(\mathbf{M}\) are the weights of \(\mathbf{G}\), \(\mathcal{L}_{\text{LM}}\) measures the language modeling loss on calibration data \(s_c\), and \(f_G(s_c; \mathbf{d}_{all})\) is the LLM forward pass affected by the current pruning-decision set $\mathbf{d}_{all}$ with Eq.\ref{eq:train} applied to all projections; \(\mathcal{L}_{\text{s}}\) enforces the sparsity constraint by encouraging the current remaining parameters \(R(\mathbf{d}_{all})\) in the decoder layers to align with the user-defined value \(r P_{\text{total}}\), \(P_{\text{total}}\) denotes the total number of parameters in all decoder layers, \(r = 1 - p\), \(p\) is the desired sparsity level, and \(\alpha\) is the sparsity constrain coefficient.

After updating \(\mathbf{G}\), the updated pruning-decision set \(\mathbf{d}_{all}\) are applied to the LoRA-weight tuning within the same ATP step. The LLM's LoRA tuning optimization objective is formulated as:
\begin{equation}
\min_{\mathbf{W}_{L}} \mathcal{J}_{L}(\mathbf{W}_{L}) := \mathcal{L}_{\text{LM}}(f_L(s_t; \mathbf{d}_{all})) + \beta\mathcal{L}_{gl},
\label{eq:loss_llm}
\end{equation}
where \(\mathbf{W}_{L}\) are the trainable LoRA weights of LLM, $f_{L}(s_t;\mathbf{d}_{all})$ is the LLM forward pass with Eq.\ref{eq:eval} applied to all projections, \(s_t\) is the training data, \(\mathcal{L}_{gl}\) promotes structured sparsity of LoRA modules, as described in Eq.5 and $\beta$ is the group lasso regularization coefficient.

ATP alternates between updating \(\mathbf{G}\) and tuning the LLM for \(T_{\text{end}}\) steps. After \(T_{\text{end}}\), the pruning-decision set \(\mathbf{d}_{all}\) are frozen, and the final tuning phase is conducted with an increased \(\beta\) to strengthen group lasso regularization, ensuring the model's sparsity structure aligns with the finalized pruning-decision set \(\mathbf{d}_{all}\).

\begin{figure*}[]
    \centering
    \includegraphics[width=\linewidth]{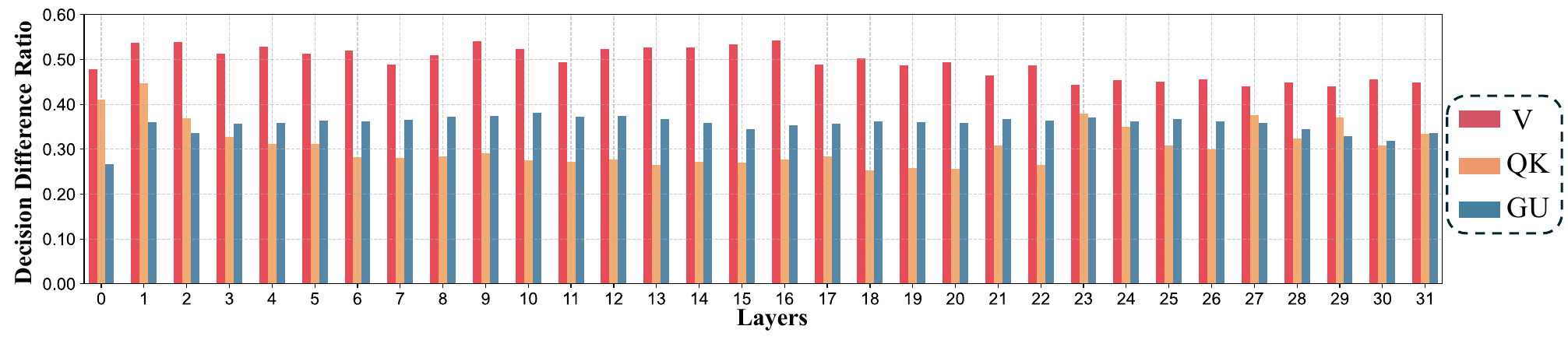} 
    \caption{Layer-wise difference ratio between the original pruning decisions generated on the pretrained model  and dynamically-adjusted decisions from ATP for LLaMA2-7B under 50\% sparsity level in \textit{HealthCare}.}
    \label{fig:evove}
    \vspace{-8pt}
\end{figure*}

\begin{figure*}[]
    \centering
    \includegraphics[width=0.9\linewidth]{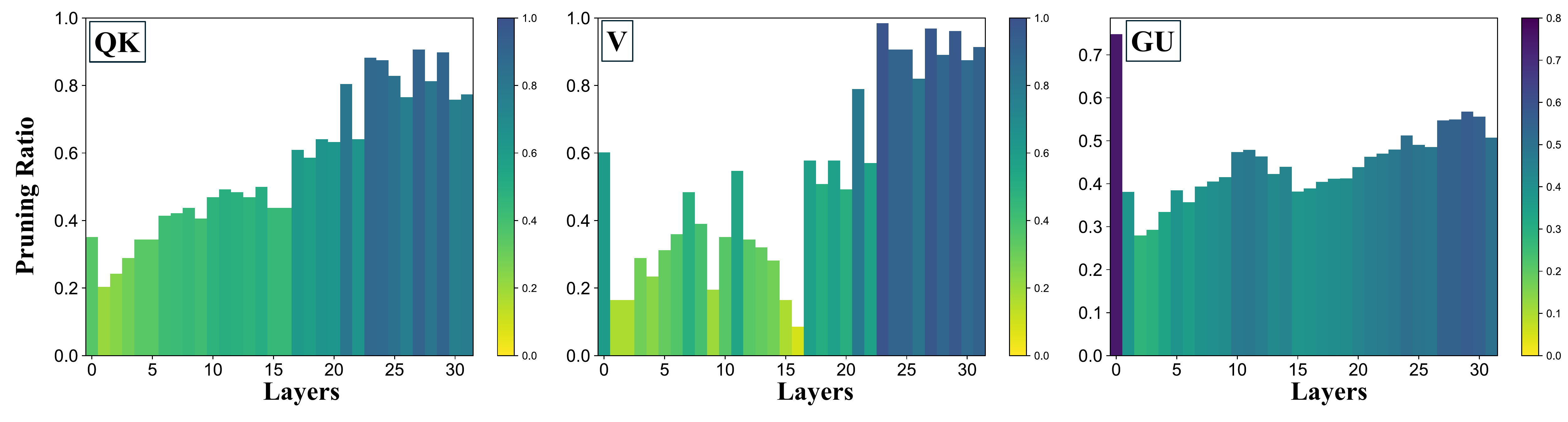} 
    \caption{Layer-wise pruning ratio according to $\mathbf{D}_{\text{\tiny QK}}$, $\mathbf{D}_{\text{\tiny V}}$ of attention and $\mathbf{D}_{\text{\tiny GU}}$ of MLP for LLaMA2-7B under 50\% sparsity level in \textit{HealthCare}. Deeper color indicates a higher pruning ratio.}
    \label{fig:sparsity}
    \vspace{-5pt}
\end{figure*}

\section{Experiment and Analysis}
\subsection{Experiment Setup}
We primarily adopt the experimental setup from \citet{zhang2024pruning} for two specific domains: \textit{HealthCare} and \textit{Legal}. 


We evaluate the performance of the pruned and domain-aligned LLM from: language modeling capability, and mult-task solving ability of natural language inference (NLI), question answering (QA), and summarization under the corresponding domain. We construct the domain-specific training and calibration datasets from MedNLI \cite{romanov2018lessons}, PubMedQA \cite{jin2019pubmedqa}, HQS \cite{abacha2019summarization} for \textit{HealthCare}, and from CaseHold \cite{zheng2021does}, BillSum \cite{kornilova2019billsum} for \textit{Legal}. 


We mainly compare ATP against two state-of-the-art structural-pruning methods LLM-Pruner \cite{ma2023llm} and SliceGPT \cite{ashkboos2023slicegpt}, where both can be categorized as two-stage pruning and tuning methods. We choose LLaMA2-7B \cite{touvron2023llama} and LLaMA3-8B \cite{dubey2024llama} from the LLaMA model family as the pretrained dense models. 


Detailed dataset construction, sample template, evaluation metrics, and hyperparameters are provided in Appendix~\ref{sec:a-2}.

\subsection{Main Results}
The overall evaluation results are shown in Tab.\ref{tab:performance_comparison}. From the table, we observe that ATP greatly outperforms LLM-Pruner and SliceGPT, two-stage pruning methods, in domain-specific language modeling and summarization capability. For language modeling, ATP  achieves the lowest perplexity under most settings, with its advantages being especially evident in the \textit{Legal} domain, where the samples are long enough to contain rich domain-specific contexts.  For summarization tasks, ATP maintains performance comparable to dense models even at higher sparsity levels while LLM-Pruner and SliceGPT struggle to capture the critical meanings and generate high-quality summaries. Although pruning impacts deterministic label prediction tasks like MedNLI and PubMedQA, ATP achieves the best results in most cases, with minor exceptions. Notably, ATP retains $75\%\sim88\%$ of the relative performance of the original dense model for LLaMA3-8B when pruning 40\% to 50\% of the parameters. For LLaMA2-7B, ATP achieves an even higher relative performance of $81\%\sim91\%$, significantly outperforming SliceGPT and LLM-Pruner in both cases.

\begin{figure*}[ht]
    \centering
    \includegraphics[width=1\linewidth]{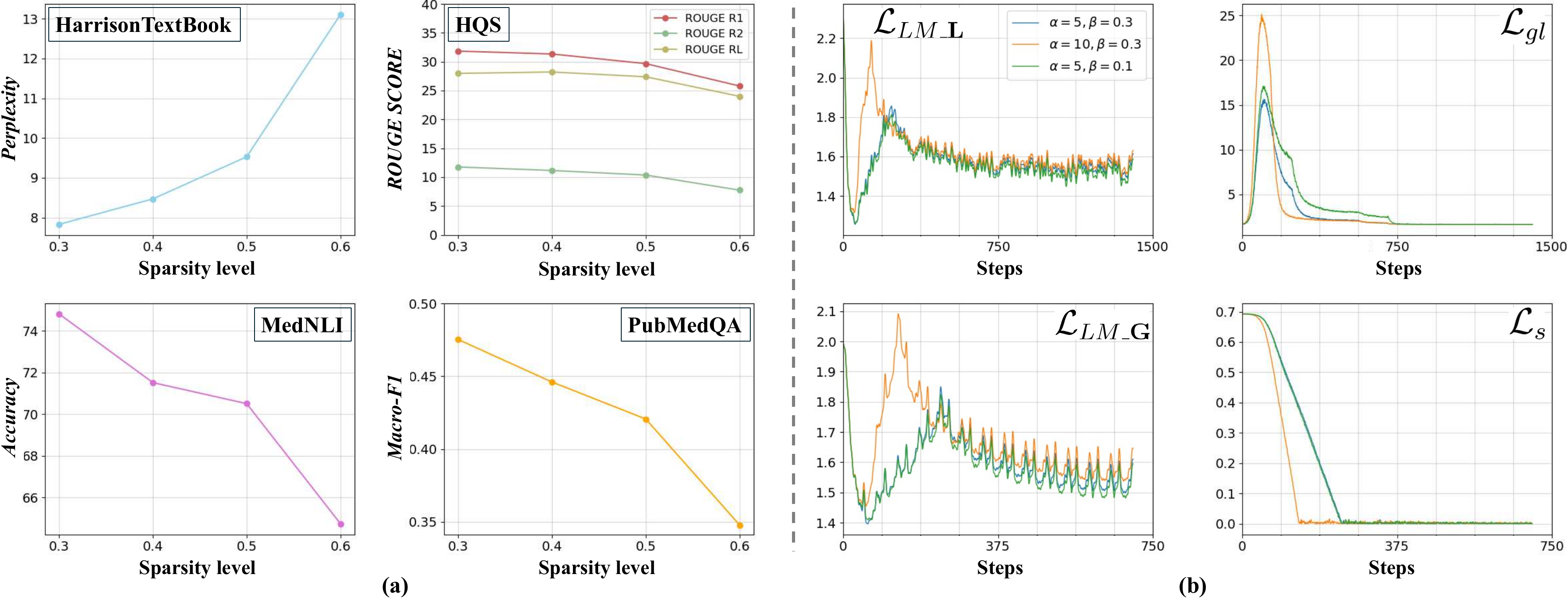} 
    \caption{(a) Task performance when changing the sparsity levels. (b) Training dynamics given different $\alpha$ and $\beta$.}
    \label{fig:analysis}
    \vspace{-5pt}
\end{figure*}

\begin{table*}[ht]
    \centering
    \small 
    \renewcommand{\arraystretch}{1.3} 
    \begin{tabular}{|p{0.2\linewidth}|p{0.75\linewidth}|}
        \hline
        
        \textbf{Question} & 
        {\footnotesize \textit{My daughter got the MMR vaccination at 1 year old, then she got a second dosage by mistake after two weeks from the first one. Is this dangerous to her? Is there anything I can do?}} \\ \hline

        \textbf{Reference} & 
        {\footnotesize \textit{Are extra doses of MMR vaccine harmful to infants?}} \\ \hline

        \textbf{LLM-Pruner} & 
        \textit{What are the effects of giving the MMR vaccine to a child?} \\ \hline
        \textbf{SliceGPT} & 
        \textit{Is there a risk from taking a second MMR vaccine?} \\ \hline
        
        \textbf{ATP (Ours)} & 
        \textit{Is second dose of MMR dangerous to child?} \\ \hline
    \end{tabular}
    \caption{Case study on summarizing a vaccination-related question. Reference is the summary provided by HQS. }
    \label{tab:case_study_mmr}
    \vspace{-5pt}
\end{table*}

\subsection{Analysis}
\textbf{Finalized Pruning Decisions.} In Fig.\ref{fig:sparsity}, we visualize the final pruning-decision set $\mathbf{d}_{all}$ generated from $\mathbf{G}$ for the LLaMA2-7B model at a 50\% sparsity level under the \textit{HealthCare} domain. The distribution of the layer-wise decision vectors $\mathbf{d}_{n}$ reveals a general trend where earlier layers are pruned less, and later layers more. This aligns with the hierarchical nature of LLMs: early layers capture general features essential for understanding the input, while later layers specialize in domain-specific details that can be pruned more aggressively. Interestingly, the first layer stands out as an outlier with a significantly higher pruning ratio than neighboring layers. We suspect this is due to the dataset consisting entirely of domain-specific tasks, where the first layer plays a relatively smaller role in domain adaptation.

\noindent\textbf{Optimal Subtructural Evolution.} To demonstrate that changes in weight importance through parameter tuning can influence pruning decisions, we train the pruning-decision generator $\mathbf{G}$ using the same initialized $\mathbf{M}$, based on the \textit{HealthCare} calibration dataset targeting at 50\% sparsity level, on the pretrained LLaMA2-7B model without LoRA fine-tuning. This allows us to identify the optimal pruning decisions on the pretrained weights only. We then compare these generated decisions, $\mathbf{d}_{all\_pt}$ with $\mathbf{d}_{all}$ from ATP, which accounts for the fine-tuning process. We mathematically evaluate the decision difference ratio via the normalized Hamming distance between two decision vectors. As shown in Fig.\ref{fig:evove}, the difference ratio of each pruning decision ranges from 20\% to even 55\%. Mathematically, this can be attributed to the changes in weight importance due to updates during the fine-tuning process. From a more abstract perspective, such decision evolution could be explained as fine-tuning enables the pruned model to recover certain capabilities more easily, while other capabilities are harder to restore. Consequently, pruning decisions are adjusted to retain weights associated with less recoverable capabilities, while pruning those that can be more easily re-established during fine-tuning. Thus, incorporating dynamically-adjusted pruning decisions together with parameter tuning is meaningful as conducted in ATP.

\noindent\textbf{Effect of sparsity level $p$.}
In Fig.\ref{fig:analysis}(a), we visualize ATP's performance on LLaMA2-7B in the \textit{HealthCare} domain under different sparsity levels, \( p = \{0.3,0.4,0.5, 0.6\} \). As expected, all metrics degrade as \( p \) increases, consistent with the general behavior of pruning. Notably, when $p$ increases from $0.3$ to $0.5$, the loss in language modeling capability and the closely related summarization performance remains minimal, while the decline in label prediction tasks such as NLI and QA stays within an acceptable range. This highlights ATP's ability to maintain performance even under significant sparsity constraints. We recommend a 50\% sparsity level with ATP as it achieves a good balance between model size and specialized performance. However, when $p$ increases to $0.6$, a sharp performance drop across all metrics occurs, suggesting that fine-tuning can no longer effectively align a heavily pruned pretrained model to the target domain under the same configuration. 


\noindent\textbf{Training Dynamics.} In Fig.\ref{fig:analysis}(b), we visualize the loss curves of ATP for \textit{HealthCare} on LLaMA2-7B targeting a 50\% sparsity level. The first row and the second row refer to the loss curves of $\mathbf{G}$ and the target LLM in Eq.\ref{eq:loss_generator} and Eq.\ref{eq:loss_llm}, respectively. The dynamics of $\mathcal{L}_{s}$, which correlates with the current sparsity level, are primarily determined by $\alpha$, while other losses are observed to be mutually affected by both $\alpha$ and $\beta$. In general, lower values of $\alpha$ and $\beta$ tend to facilitate more stabilized training and better language modeling on the calibration dataset but result in a slower approach toward the desired sparsity level. 
Moreover, through joint analysis of $\mathcal{L}_{gl}$ and $\mathcal{L}_{LM\_G}$, we observe that due to the small size of the calibration dataset for $\mathbf{G}$ training, a small portion of the pruning decisions may change depending on the calibration samples. This is because it is difficult to find the theoretically global optimal decision for every sample. Thus, freezing $\mathbf{G}$ and increasing $\beta$ are crucial to facilitate the coverage of the pruned portions of LoRA weights towards zero and to achieve the desired structural sparsity after the ATP process.


\noindent\textbf{Case Study.} We perform a summarization case study on a vaccine-related \textit{HealthCare} question using LLaMA2-7B with  50\% sparsity level across all methods. As shown in Tab.\ref{tab:case_study_mmr}, the reference summary highlights two critical points: \textit{extra dose} and \textit{infant}. However, SliceGPT overlooks the \textit{infants}, providing a more generic response, while LLM-Pruner fails to capture the concern of the \textit{extra dose}. In contrast, ATP successfully incorporates both key points, demonstrating the stronger language modeling capability over other methods. More case study examples are given in the Appendix~\ref{sec:a-3}.

\section{Conclusion}
In this work, we introduce ATP, a novel one-stage structural pruning and tuning method for domain-specific LLM compression. More specifically, ATP integrates dynamically adjusted pruning decisions with sparsity-regularized fine-tuning, leveraging a LoRA-aware design to effectively address the challenges posed by limited tuning data in specialized domains. Extensive experiment results demonstrate that ATP greatly outperforms the conventional two-stage methods in language modeling and multi-task solving capabilities in specific domains under diverse settings. We believe such a unified one-stage framework and the related findings would open new possibilities for domain-specific LLM pruning.

\clearpage
\section{Limitations}
While ATP demonstrates its effectiveness in domain-specific LLM compression, it still has some common issues observed in existing pruning techniques. For example, in summarization tasks, for a small subset of input samples, the pruned model may occasionally get stuck repeating certain words or short phrases although the likelihood of this issue is reduced compared to other methods. Furthermore, achieving extremely high sparsity levels (i.e., $p \geq 0.6$) remains a significant challenge as we observe a relatively substantial performance drop when transitioning from $p=0.5$ to $p=0.6$, indicating that the aggressive removal of parameters at such sparsity levels can disproportionately affect the model’s ability to retain and recover critical knowledge. Such degradation remains difficult to mitigate quickly, suggesting that additional strategies are required to ensure the model's resilience under extreme sparsity constraints.

Future work may aim to address these limitations to unlock the full potential of our proposed ATP method for practical and scalable deployment in domain-specific applications.

\section{Ethical Considerations}
In this paper, we focus on pruning and fine-tuning domain-specific LLMs. Data samples for specialized domains such as healthcare, legal, medical, and finance often have the potential to contain sensitive or private information, raising critical ethical and privacy concerns. Here, we make it clear that all training and testing samples used in our experiments for both the \textit{HealthCare} and \textit{Legal} domains are drafted exclusively from public open-source datasets. These datasets have undergone rigorous pre-processing before being made publicly to eliminate any personal or sensitive information to minimize the risk of private data leakage. For example, in the HQS dataset, a raw sample would look like: \textit{“My name is [NAME] from [LOCATION], my 10-year-old boy has Duchenne muscular dystrophy…”} with all personal information removed.

\bibliography{latex/custom, latex/arxiv} 

\clearpage
\appendix

\section{Appendix}

\subsection{Pruning-Decision Generator $\mathbf{G}$}~\label{sec:a-1}
In general, $\mathbf{G}$ does not take external inputs to generate the pruning-decision vectors. The adjustment of pruning-decision vectors is achieved by updating the trainable parameters in $\mathcal{G}$.

\subsubsection*{Network Architecture}

In our implementation, the pruning-decision generator $\mathbf{G}$, as shown in Tab.\ref{tab:generator}, sequentially consists of the following components:

\textbf{Self input.}  The input of $\mathbf{G}$ is a frozen orthogonally-initialized \textit{nn.Parameter} with shape $(N, 64)$, where $N$ is the total number of decoder layers of the target LLM.

\textbf{Transformer Encoder.}  The input is firstly fed into 2 sequential Transformer encoder blocks~\cite{vaswani2017attention}. Each block uses a multi-head self-attention mechanism with 4 attention heads and a feed-forward network with an intermediate dimension of $256$ and the ReLU activation function~\cite{krizhevsky2012imagenet}. The encoders process the input into a tensor of shape $(N, 64)$, representing the layer-wise intermediate representations.

\textbf{Layer Normalization.}  
A layer normalization~\cite{Ba2016LayerN} module is then applied to the intermediate representation, ensuring stabilized training subsequent computations. The output of this step maintains the shape $(N, 64)$.

\textbf{Layer-Wise Decision Projection.}  
The output of the previous LayerNorm is then projected to the desired length, counting for all pruning decisions for a single decoder layer in the target LLM. Specifically, the projected dimension $ = d_{head} * 2 + d_{int}$, where $d_{head}$ is the head-wise attention dimension and $d_{int}$ is the intermediate projection dimension in MLP. For example, for LLaMA2-7B, $d_{head} = 128$, $d_{int} = 11008$, thus the projected dimension is $128 * 2 + 11008 = 11264$, counting for total the pruning decisions for $\mathbf{D}_{QK}$, $\mathbf{D}_{V}$ and $\mathbf{D}_{GU}$ as mentioned in the paper. The output shape of the decision projection is $(N, d_{head} * 2 + d_{int})$.

\textbf{Gumbel-Sigmoid Sampling and Binary Mask Conversion via STE.}  
Then the projected output is sampled via Gumble-Sigmoid with sampling temperature $T=0.4$ and offset base $=3$ to approximate binomial distribution for each decision. We consider the decision elements in the output tensor after Gumbel-Sigmoid Sampling with the shape $(N, d_{head} * 2 + d_{int})$ as \textit{`soft'} decisions.

\begin{table}[]
    \caption{The architecture of the Generator. $H$ represents the number of heads.}
      \centering
      \resizebox{1.0\columnwidth}{!}{
        \begin{tabular}{l}
            \toprule
                  Input (N, 64)\\
            \midrule
                  MHA($H=4$)$\rightarrow$Add Residual$\rightarrow$ FFN(64$\rightarrow$256$\rightarrow$ReLU$\rightarrow$LayerNorm$\rightarrow$64)$\rightarrow$Add Residual\\
                  MHA($H=4$)$\rightarrow$Add Residual$\rightarrow$ FFN(64$\rightarrow$256$\rightarrow$ReLU$\rightarrow$LayerNorm$\rightarrow$64)$\rightarrow$Add Residual\\
                  \midrule
            $\textrm{Projection}_n$(64, $d_{head} * 2 + d_{int}$)$\rightarrow$ ${\mathbf{d}_{soft}}_n$, $n=1,\cdots, N$\\
            \bottomrule
              \\
          \end{tabular}
          }
          \label{tab:generator}
\end{table}

To achieve the actual binary `$0$' or `$1$' decision, we further round those soft decisions $\geq 0.5$ to `$1$' and $<0.5$ to `$0$' together with STE~\cite{bengio2013estimating} to maintain differentiability and achieve the final \textit{`hard'} decisions. The final output of $G$ is a tensor consisting of `$0$' and `$1$' with the shape $(N, d_{head} * 2 + d_{int})$, serving as the pruning-decision vectors for $N$ decoder layers of the target LLM. More specifically, given the outputs of the Generator, we can calculate the binary mask as:
\begin{equation}~\label{eq:gs}
    \mathbf{d}_n = \text{round}(\text{sigmoid}(({\mathbf{d}_{soft}}_n + g + b)/T)),
\end{equation}
where $b$ is the offset base, $T$ is the temperature hyperparameter, and $g\in \text{Gumbel}(0,1)$, $\text{Gumbel}$ is the Gumbel distribution, and $\text{round}(\cdot)$ rounds the input to the nearest integer.

\subsection{Detailed Experiment Setup}~\label{sec:a-2}
\subsubsection*{\textbf{Dataset Construction}}
We construct the domain-specific training datasets following the same settings of D-Pruner \cite{zhang2024pruning}. 

For the \textit{HealthCare} domain, we create the fine-tuning dataset $\mathcal{D}_{t}$ by combining MedNLI, PubMedQA, and HealthQuestionSummary (HQS) in a ratio of 7:7:1, resulting in a total of 15,000 samples. For the \textit{Legal} domain, we select data from CaseHold and BillSum in a ratio of 13:2, also totaling 15,000 samples. All training samples are obtained by selecting the first $n$ samples from the `train' split of each respective dataset.

The calibration dataset $\mathcal{D}_{c}$ for both domains contains 1,000 samples, with a ratio of 1:3:1 from MedNLI, PubMedQA, and HQS for the \textit{HealthCare} domain, and 1:1 from CaseHold and BillSum for the \textit{Legal} domain. Inspired by the concept that general weight importance obtained from open-domain datasets enhances the model's adaptability and generalization across multiple tasks, we further extract the first 300 samples from the C4 \cite{raffel2020exploring} dataset and append them to each of the two domain-specific calibration datasets, resulting in a total of 1,300 samples for each domain-specific calibration dataset.

It is worth mentioning that we \textbf{DO NOT} include any samples from HarrisionTextBook (a widely recognized and authoritative medical textbook for \textit{HealthCare}) and MultiLegalPile \cite{niklaus2023multilegalpile} (a large dataset consisting various documents relevant to law and legal proceedings for \textit{Legal}) in either $\mathcal{D}_t$ or $\mathcal{D}_c$.

For \textit{HealthCare} performance evaluation, we choose the first 300 paragraphs from HarrisionTextBook, 
the entire 1422 test samples in MedNLI `test' split, the first 500 test samples in PubMedQA `test' split, and the entire 100 test samples in HQS `test' split. For \textit{Legal} performance evaluation, we choose the first 300 samples in the `en\_legislation\_US' split of MultiLegalPile, and the first 200 test samples in BillSum `test' split.

\subsubsection*{\textbf{Sample Template}}
D-Pruner \cite{zhang2024pruning} mentions that all the samples are formulated following Alpaca template \cite{Bommasani2021FoundationModels},
thus we follow this for all the samples except for those from HarrisonTextBook and MultilegalPile because they are for perplexity evaluation only. For better understanding, we provide our formulated template for MedNLI as an example:

"\textit{Below is an instruction that describes a task related to HealthCare, paired with an input that provides further context. Write a response that appropriately completes the request.}

\textit{Instruction: Determine the relationship between the HealthCare Premise and the Hypothesis from `entailment', `contradiction', `neutral'.}

\textit{Input: Premise: `\{sentence1\}', Hypothesis: `\{sentence2\}.}

\textit{Response: Their relationship is \{label\}}. ",

where sentence1, sentence2, and label are the corresponding Premise, Hypothesis, and ground-truth labels extracted from each drafted sample from MedNLI. Samples from other datasets are also formulated in a similar way.

\subsubsection*{\textbf{Evaluation Metrics}} 
We evaluate the pruned models' language modeling (linguistic) capabilities using perplexity scores on HarrisonTextBook and MultiLegalPile, corresponding to the \textit{HealthCare} and \textit{Legal} domains, respectively. The natural language inference (NLI) abilities are assessed through prediction accuracy on MedNLI. For question answering (QA), we evaluate the models using Macro-F1 scores on PubMedQA since the data distribution in PubMedQA is super biased and the majority of the answers are \textit{`yes'}. Summarization capabilities are measured using ROUGE scores \cite{lin2004rouge} on HQS and BillSum for \textit{HealthCare} and \textit{Legal} domain. Considering the randomness in next-token sampling, we conduct each evaluation 3 times except for perplexity evaluation (as perplexity calculation does not involve token sampling), to report the average score of it. 

To better demonstrate the overall performance of the pruned model compared to the dense model, we formulate the Relative Performance of a pruned model as follows: For metrics within a domain, 
\begin{equation}~\label{eq:rel-pef}
    \text{Relative Performance} = \frac{1}{n} \sum_{i=1}^{n} \frac{\text{Score}_{p,i}}{\text{Score}_{d,i}}.
\end{equation}
Where $\text{Score}_{p,i}$ and $\text{Score}_{d,i}$ are scores for the pruned and the dense model for $i$th task separately. For summarization metrics such as ROUGE R1, R2, and RL, we define the performance ratio between the pruned model and the dense model as: $\frac{\text{Summary}_p}{\text{Summary}_d}=\frac{1}{3}(\frac{\text{R1}_{p}}{\text{R1}_{d}} + \frac{\text{R2}_{p}}{\text{R2}_{d}}+\frac{\text{RL}_{p}}{\text{RL}_{d}})$. This ratio, $\frac{\text{Summary}_p}{\text{Summary}_d}$, is then inserted into Eq.~\ref{eq:rel-pef} and averaged with scores from other tasks.. 

\begin{table*}[t]
    \centering
    \small 
    \renewcommand{\arraystretch}{1.3} 
    \begin{tabular}{|p{0.2\linewidth}|p{0.75\linewidth}|}
        \hline
        
        \textbf{Question} & 
        {\footnotesize \textit{"I have been suffering from digestive problems for 30 years. It has recently flared up bad and I am looking for ideas to try to heal. I was diagnosed with IBS at 13.  I was curious about the servings in the C-IBS Formulation. How's much of each in mgs? How many times a day? I want to take a more holistic approach as I'm going to be 42 and I am tired of suffering and the side effects of my anti-nausea meds that left me with an eye tic."}} \\ \hline

        \textbf{Reference} & 
        {\footnotesize \textit{What is in the C-IBS formula and how should it be used? How do you treat IBS naturally?}} \\ \hline

        \textbf{SliceGPT} & 
        \textit{What is the diet for digestive problems and how many drinkings per day are required for IBS?} \\ \hline

        \textbf{LLM-Pruner} & 
        \textit{How much C-IBIS should I take for daily digestive complaints?} \\ \hline

        \textbf{ATP (Ours)} & 
        \textit{What are the ingredients of C-IBS and how much does each serve?} \\ \hline
    \end{tabular}
    \caption{Case study on summarizing a question related to digestive problems in the \textit{HealthCare} domain.}
    \label{tab:case_study_digestive}
\end{table*}

\begin{table*}[t]
    \centering
    \small 
    \renewcommand{\arraystretch}{1.3} 
    \begin{tabular}{|p{0.2\linewidth}|p{0.75\linewidth}|}
        \hline

        \textbf{Bill} & 
        {\footnotesize \textit{"Farm to School Improvements Act of 2010": The Secretary shall provide assistance, through competitive matching grants and technical assistance, to eligible entities for farm to school programs that improve access to local foods...}} \\ \hline

        \textbf{Reference} & 
        {\footnotesize \textit{Farm to School Improvements Act of 2010 - Amends the Richard B. Russell National School Lunch Act to direct the Secretary of Agriculture to provide competitive matching grants to schools, nonprofit organizations, and other able entities for farm to school programs that improve the access of school lunch and breakfast program participants to local foods. Provides that each grant may include an implementation grant, training and technical assistance grant, and planning grant. Requires farm to school programs to be designed to: (1) improve the nutritional health and well being of children, (2) procure healthy local foods from small and medium-sized farms, (3) support experiential nutrition education by involving school children in farm and garden-based agricultural education activities, (4) commit public and private community stakeholders to the sustained success of such programs, and (5) increase farmers' income by facilitating their access to institutional markets. Directs the Secretary to provide grant recipients with technical assistance that includes sharing information, best practices, research, and data on existing farm to school programs.}} \\ \hline

        \textbf{SliceGPT} & 
        \textit{Amends the Richard B. Russell School Lunch Act to provide assistance to a community for agriculture-based farm to school programs to improve agricultural health and well-being.} \\ \hline

        \textbf{LLM-Pruner} & 
        \textit{Directs the Secretary of Agriculture to make grants to public schools and non-profit organizations, through such assistance as the Secretary may require, to provide food that is nutritionally appropriate, locally produced, and available to students in local schools.} \\ \hline

        \textbf{ATP (Ours)} & 
        \textit{Farm to School Improvement Program. Authorizes the Secretary of Agriculture (USDA) to provide assistance, through competitive matching grants and technical assistance, to eligible entities for farm to school programs that improve access to local foods in schools and institutions participating in programs under the Richard B. Russell National School Lunch Act. Establishes a priority for the grant awarding authority to assist recipients in promoting local food production and to ensure geographical diversity in the awarding of grant funds. Requires the Secretary to form review panels to evaluate proposals based on the extent to which the proposed program improves the nutritional health and well being of children with specific evaluation criteria.} \\ \hline

    \end{tabular}
    \caption{Case study on summarizing the "\textit{Farm to School Improvements Act of 2010}" in the \textit{Legal} domain.}
    \label{tab:case_study_legal}
\end{table*}

\subsubsection*{\textbf{Baselines}}

To ensure a fair comparison, both LLM-Pruner and SliceGPT use the same calibration dataset for stage 1 (pruning), the same training dataset for stage 2 (post-pruning tuning), and the same LoRA tuning hyperparameters as ATP. To achieve this, we handcraft the dataset formulation part of their provided post-pruning tuning Python scripts to integrate into our formulated domain-specific datasets as their original scripts only support general datasets from Huggingface. 

\subsubsection*{\textbf{Training Details}}
We conduct the training for every method on 8$\times$ NVIDIA A100s with 80G cuda memory. For every method, no samples were truncated.

\textbf{LLM Loss Modeling.} Unlike D-Pruner \cite{zhang2024pruning}, which models the next-token prediction loss solely on the \textit{reponse}, we adopt the tuning approach from \citet{shi2024instruction} in our implementation, where the next-token prediction loss is computed on the entire sentence (\textit{instruction} + \textit{response}). We consider this adjustment necessary because we find a large portion of \textit{response} in the formulated domain-specific datasets are overly simplistic, often limited to templated responses such as ``yes", ``no" or ``maybe". In contrast, the \textit{instruction} contains valuable domain-specific knowledge which is important for domain alignment. We also apply such LLM Loss modeling for the compared baselines during their fine-tuning stage for fairness comparison.

\textbf{Training Hyperparameters.} For ATP and compared baselines, we adopt the following same hyperparameters to achieve comparison fairness: we set LoRA rank $r=8$, learning rate for LoRA modules $=1e^{-4}$, total number of epochs $= 3$, local mini-batch size $= 4$ for \textit{HealthCare} and $1$ for \textit{Legal} because some samples within \textit{legal}-specific dataset are extremely long, equivalent global mini-batch size $=32$ and 8, equivalent total step $T = 1406$ and $5624$. For ATP, besides those shared hyperparameters, we set the learning rate for pruning-decision generator $\mathbf{G}$ as $5e^{-4}$, end of $\mathbf{G}$ training $T_{end} = \frac{T}{2}$, coefficient of pruning-decision sparity constrain $\alpha = 5$, coefficient group\_lasso\_regularization $\beta = 0.3$, and increase $\beta$ to $100*\beta$ after $T_{end}$ to facilitate structural-sparsity convergence. We use AdamW(0.9, 0.999) for both LoRA weights optimization and $\mathbf{G}$ weights optimization.

For next-token sampling, for all the methods, we use the same configuration of top\_k = 50, top\_p = 0.9, and temperature = 0.9 to ensure evaluation fairness.

\subsection{More Case Study}~\label{sec:a-3}
We provide additional case studies on summarization tasks on both domains as shown in Tab.\ref{tab:case_study_digestive} and Tab.\ref{tab:case_study_legal} based on LLaMA2-7B under 50\% sparsity level. The results demonstrate that our method produces summaries that better capture the key points of the input with better domain-specific language consistency, while SliceGPT and LLM-Pruner relatively miss critical details or fail to align with the specialized terminology required for each domain.

\subsection{Reported Results of LLM-Pruner}
D-Pruner \cite{zhang2024pruning} also reports the results of LLM-Pruner for \textit{HealthCare} and \textit{Legal} on LLaMA2-7B under 50\% sparsity level. We find a noticeable difference between our reported results and the D-Pruner reported. For instance, \citet{zhang2024pruning} reports perplexity scores of 44.56 and 215.13 for \textit{HealthCare} and \textit{Legal}, respectively, whereas our reported scores for LLM-Pruner are 13.67 and 7.97. We suspect this is mainly caused by different LLM loss modeling as mentioned before because we model the loss on the entire samples. Besides, D-Pruner has not released their fine-tuning scripts yet and we think different hyperparameter settings may also account for such divergence. We gently consider our training settings could achieve a more stabilized fine-tuning process and results.

\subsection{Repeated pattern in $\mathcal{L}_{LM\_\mathbf{G}}$}
As shown in the loss curve of LLM Loss on the calibration dataset $\mathcal{L}_{LM\_\mathbf{G}}$ for pruning-decision generator $\mathbf{G}$ training in Fig.\ref{fig:analysis}(b), a repeated pattern could be observed. The reason is that the calibration dataset contains only 1.3K samples, while the training dataset consists of 15K samples for both the \textit{HealthCare} and \textit{Legal} domains. To simplify the implementation of the training scripts, we currently construct the dataloader for the calibration dataset as an iterable object. This is done using \textit{itertools.cycle}, as shown: \textit{mini\_batch\_calibration = itertools.cycle(calibration\_dataloader)}. As a result, there is no randomness in sampling from the calibration dataset. Such repeated pattern is caused by periodic cycling through the same order of mini-batches of calibration data. We would consider in our following research whether it is necessary to randomly draft mini-batches from the calibration dataset for pruning-decision generator $\mathbf{G}$ training.


\subsection{ATP for Full-Parameter Fine-tuning}
Even though ATP is a LoRA-oriented method, it could be theoretically extended towards full-parameter fine-tuning with minimal effort.

Without inserted LoRA modules, We could simply rewrite Eq.\ref{eq:eval} and Eq.\ref{eq:train} as:
\begin{equation*}
    f_{G}(\mathbf{X}) = \mathbf{X}\mathbf{W}\mathbf{D}
\label{eq:eval_full}
\end{equation*}
and 
\begin{equation*}
    f_{L}(\mathbf{X}) = \mathbf{X}\mathbf{W}.
\label{eq:train_full}
\end{equation*}
and group lasso regularization for achieving structural sparsity could be straightforwardly applied on the row of $(\mathbf{I} - \mathbf{D}_{prev})\mathbf{  W}$ and columns of $\mathbf{W}(\mathbf{I}-\mathbf{D})$. After training is done, desired pruned rows and columns can be directly removed from $\mathbf{W}$. However, due to the limited amount of domain-specific datasets, we would still recommend using LoRA as the tuning method. We would like to explore the proper application scenarios of ATP for full-parameter fine-tuning in future research.

\subsection{Discuss on the Artifacts}
The licenses for various models and datasets are as follows: \textbf{LLaMA2 and LLaMA3}: Licensed under the LLaMA 2 Community License and META LLaMA 3 Community License; \textbf{PubMedQA}: Licensed under MIT License; \textbf{HQS}: Licensed under Apache License 2.0; \textbf{MedNLI}: Licensed under The PhysioNet Credentialed Health Data License
Version 1.5.0; \textbf{CaseHold}: Licensed under Apache License 2.0.

\end{document}